\PassOptionsToPackage{table,xcdraw}{xcolor}

\documentclass[9pt,sigconf,letterpaper]{acmart}

\usepackage[english]{babel}
\usepackage{blindtext}

\usepackage{tikz}

\usepackage{paralist}

\usepackage{soul}
\renewcommand\footnotetextcopyrightpermission[1]{} 
\setcopyright{acmcopyright}
\setcopyright{acmlicensed}

\usepackage{placeins}

\newcommand{\eat}[1]{}

\newcommand{\etal}{{\it et al.}\xspace}

\acmDOI{}

\acmISBN{}

\acmPrice{}

\usepackage{tabularx}
\newcolumntype{L}[1]{>{\raggedright\arraybackslash}p{#1}}
\newcolumntype{C}[1]{>{\centering\arraybackslash}p{#1}}
\newcolumntype{R}[1]{>{\raggedleft\arraybackslash}p{#1}}

\usepackage{algorithm}
\usepackage[noend]{algorithmic}

\usepackage{enumitem}
\setlist{nolistsep}
\usepackage{graphicx}
\usepackage{epstopdf}
\usepackage{grffile}
\usepackage{amsmath}
\usepackage{multicol,multirow}
\usepackage{rotating,url}
\usepackage{enumitem}
\usepackage{color}
\usepackage{subfigure}
\usepackage{xspace}
\usepackage{ifpdf}
\usepackage{mdwlist}
\usepackage{colortbl}
\usepackage{caption}
\usepackage{booktabs}
\usepackage{multirow}
\usepackage{enumitem}
\usepackage{xfrac}
\usepackage{afterpage}
\usepackage{pdflscape}
\usepackage{longtable}
\usepackage{hhline}
\usepackage{lscape}

\newlength{\oldtextfloatsep}

\usepackage{enumitem}
\usepackage{mdwlist}
\setlist{nolistsep}
\setlist[description]{noitemsep,topsep=0pt,parsep=0pt,partopsep=0pt,leftmargin=0pt}
\usepackage{pifont}

\usepackage{everypage}

\newcommand{\todo}[1]{\textcolor{red}{TODO: \emph{#1}}}
\newcommand{\tocut}[1]{\textcolor{blue}{\emph{#1}}}
\newcommand{\oldnotes}[1]{\textcolor{gray}{#1}}
\newcommand{\amm}[1]{\textcolor{magenta}{AMM: \emph{#1}}}
\newcommand{\djd}[1]{\textcolor{teal}{DJD: \emph{#1}}}
\newcommand{\rk}[1]{\textcolor{purple}{RK: \emph{#1}}}
\newcommand{\drc}[1]{\textcolor{olive}{[[DC: \emph{#1}]]}}
\eat{
\newcommand{\todo}[1]{}
\newcommand{\tocut}[1]{}
\newcommand{\oldnotes}[1]{}
\newcommand{\amm}[1]{}
\newcommand{\djd}[1]{}
\newcommand{\rk}[1]{}
\newcommand{\drc}[1]{}
}

\long\def\comment#1{}

\newcommand{\Lpagenumber}{\ifdim\textwidth=\linewidth\else\bgroup
  \dimendef\margin=0 
  \ifodd\value{page}\margin=\oddsidemargin
  \else\margin=\evensidemargin
  \fi
  \raisebox{\dimexpr -\topmargin-\headheight-\headsep-0.5\linewidth}[0pt][0pt]{%
    \rlap{\hspace{\dimexpr \margin+\textheight+\footskip}%
    \llap{\rotatebox{90}{\thepage}}}}%
\egroup\fi}
\AddEverypageHook{\Lpagenumber}%

\begin{document}


\clubpenalty=10000 
\widowpenalty = 10000

\title[Rapid IoT Device Identification at the Edge]{Rapid IoT Device Identification at the Edge}


\author{Oliver Thompson}
\affiliation{
  \institution{Imperial College London}
\country{}
}
\author{Anna Maria Mandalari}
\affiliation{
  \institution{Imperial College London}
\country{}  
}
\author{Hamed Haddadi}
\affiliation{
  \institution{Imperial College London}
\country{}
}

\renewcommand{\shortauthors}{Thompson et al.} 

\acmYear{2021}\copyrightyear{2021}
\setcopyright{acmcopyright}
\acmConference[DistributedML '21]{2nd International Workshop on Distributed Machine Learning DistributedML 2021}{December 7, 2021}{Virtual Event, Germany}
\acmBooktitle{2nd International Workshop on Distributed Machine Learning DistributedML 2021 (DistributedML '21), December 7, 2021, Virtual Event, Germany}
\acmPrice{15.00}
\acmDOI{10.1145/3488659.3493777}
\acmISBN{978-1-4503-9134-4/21/12}

\begin{abstract}


Consumer Internet of Things (IoT) devices are increasingly common in everyday homes, from smart speakers to security cameras. Along with their benefits come potential privacy and security threats. To limit these threats we must implement solutions to filter IoT traffic at the edge. 
To this end the identification of the IoT device is the first natural step. 

In this paper we demonstrate a novel method of rapid IoT device identification that uses neural networks trained on device DNS traffic that can be captured from a DNS server on the local network. The method identifies devices by fitting a model to the first seconds of DNS second-level-domain traffic following their first connection. Since security and privacy threat detection often operate at a device specific level, rapid identification allows these strategies to be implemented immediately.
Through a total of 51,000 rigorous automated experiments, we classify 30 consumer IoT devices from 27 different manufacturers with 82\% and 93\% accuracy for product type and device manufacturers respectively.

\end{abstract}



\maketitle

\section{Introduction}\label{sec:intro}

The consumer Internet of Things (IoT) space has experienced a significant rise in popularity in recent years. From smart speakers, to baby monitors, these devices are becoming increasingly common in households~\cite{IoT-stats}.
Since there are no strict compliance and regulations in this ecosystem, IoT malware, botnets, and device abuse (e.g., external access leading to domestic abuse) is increasingly becoming a recurring and major security and privacy issue~\cite{ren-imc19, 10.1145/3319535.3354198, varmarken2020tv}. On the other hand, due to the presence of several middleboxes, gateways and traffic sampling at ISPs; it is practically impossible to identify, detect, and isolate the misbehaving devices or households. This means devices at the edge are best positioned to defend against these attacks. 

IoT devices would benefit from automated management of these privacy and security threats~\cite{8960276,mandalari2021blocking}. The first natural step is to automate the identification of the device at the edge.
There have been several solutions proposed for IoT device identification~\cite{AUDI-IoT-JSAC, 8440758, 9097761, 9346251, 8026581, kolcun2019case}.  
Those approaches rely on training machine learning models offline or in a cloud environment, using network traffic. However, the training and validation of these models is achieved using a list of features of a particular set of devices, over a long time period.
Moreover, these approaches rely on continuous and complete packets capture and data collection, which is not feasible on a device at the edge which likely has limited computational resources.

In this paper we propose a novel method of rapid IoT device identification using neural networks trained on device DNS log traffic that can be captured from a DNS server on the local network. 
Our method is able to accurately identify the device type from the first few seconds of traffic after the device is connected.
The method identifies devices by fitting a model to the first minute of DNS second-level-domains traffic following their first connection. 

It is important that device identification occurs rapidly so that device-specific privacy and security threats can be mitigated immediately following a device's first connection. By only considering the traffic following the first connection we are able to fingerprint the product's traffic in a more repeatable way; as the user will not have much time to introduce variation into the traffic by using the product, and there may be certain consistent traffic behavior triggered by the first time connection.

Through a total of 51,000 rigorous automated controlled experiments, collected during different periods of time from 30 IoT devices, we characterize the minimum amount of time necessary to identify the device.
Results demonstrate that the model reaches maximum accuracy when trained on only 30 seconds of data after the device is connected, and can therefore perform accurate classification very quickly. 
We also demonstrate that the model retains high accuracy when tested with data collected several days after the training period. 
At product level granularity the accuracy and macro f1 score is 85\% and 0.87 respectively, when classifying the devices manufacturer rather than product type, the accuracy and f1 score were 95\% and 0.91. 

We sample the design space of neural networks and compare 1,800 model configurations to determine the best neural network architecture. we ascertain that a model with an input dimension of 32 and 2 hidden layers is highly accurate on the training data and can retain this accuracy when tested on unseen data collected several days after the training period. 
Unlike similar approaches, this method does not require full packet capture, which is computationally expensive and may pose privacy concerns.

Our main contributions are as follows:

\begin{itemize}
\item We develop a methodology for identifying IoT devices using the first  30 seconds of DNS traffic.
\item We show that it is possible to detect device product with an accuracy of 82\%, device manufacturers can be predicted with an accuracy of 93\%.
\item We demonstrate that accuracy is retained for at least one week following training data collection.
\end{itemize}

\section{Methodology}\label{sec:method}

In this section we cover the data collection and experiment methodology. We describe the testbed we use for conducting the experiments, the IoT devices under test, and the neural network architecture we use to identify the devices.

\subsection{Testbed and IoT Devices}
Our methodology rely on a controlled environment for testing IoT devices.
Our testbed consists of: 
\begin{itemize}
\item A \emph{router} that offers IP connectivity to the IoT devices under test, and the ability to capture network traffic for each device;
\item A \emph{DNS server} under our control, that serves as a proxy for the ISP’s DNS server.
\item \emph{Smart-plugs} which can be turned on and off programmatically;
\item A set of \emph{support scripts} to control the smart plugs to turn on and off an IoT device automatically. 
\end{itemize}
The support scripts are used to systematically switch the IoT devices off and on through the smart-plugs and collect the device's first traffic following reboot as a PCAP file into a structured directory.

Table~\ref{table:devices} describes the IoT devices we use in our experiments, by category. We consider in total 30 IoT devices from 27 manufacturers, chosen for the popularity and prevalence in homes.

\begin{table}[!bpt]
  \captionsetup{skip=0.2em}
	\centering
	\rowcolors{3}{gray!10}{white}
		\resizebox{1.0\columnwidth}{!}{	
		\Huge
\begin{tabularx}{\textwidth}{l|X}
			\textbf{Category}& \textbf{Device Name}
			\\ \hline
\emph{Audio} & 
Echo~Spot, Echo~Plus, Google~Home
\\ \hline
\emph{Camera} & 
Blink~Cam, Bosiwo~Cam, Wansview~Cam, Yi~Cam
\\ \hline
\emph{Home Automation} &
Anova~Sousvide, Cosori~Cooker, Gosund~Bulb, Govee~Strip, Honeywell~T-stat, Levoit~Humidifier, Magichome~Strip, Meross~Door Opener, Netatmo~Weather, Smarter~Coffee~Machine, Smartlife~Remote, TP-Link~Bulb, TP-Link~Plug, Wemo~Plug
\\ \hline
\emph{Smart Hubs} &  
Insteon, Lightify, Philips~Hue, Sengled, Smartthings, SwitchBot, Xiaomi 
\\ \hline
\emph{Video} &
Fire~TV, Samsung~TV 	
\\ \hline
\end{tabularx}
}
\caption{IoT devices under test.} 
\label{table:devices}
\end{table}

\subsection{Dataset Generation}
We perform 51,000 on-off experiments. Each experiment turns the devices on and off 100 times every two minutes, and this process is scheduled to run every 12 hours during one week.
We use Python 3's Scapy library~\cite{scapy} to parse each PCAP file by identifying the timestamp of the DHCP discover packet and creating a list of all outbound DNS queries and their respective timestamps, following this DHCP discovery. We then save the Python objects containing the DNS data, alongside their source PCAP file.

In order to use the DNS traffic as input to a predictive model it is necessary to reduce the wide range of possible URLs to a set of discrete buckets. To achieve this, we pass the SLD (second-level domain) to a hash function, the result of the hash function is reduced into $h$ buckets using the modulo operator, where $h$ is the hash resolution and the hashing function is Python 3's built in hash function. See equation~\ref{eq:hash-res}.

\begin{equation} \label{eq:hash-res}
    \begin{split}
        &URL =\ 'time1.google.com' \\
        &SLD =\ 'google' \\
        &datapoint = hash(SLD)\ \%\ h
    \end{split}
\end{equation}

We test different values of hash resolutions between $4$ and $64$. A higher hash resolution reduces the chances of unrelated SLDs colliding, and therefore results in smaller information loss. However, a lower hash resolution reduces the complexity of the model and size of the dataset. 

We choose the SLD as the input to the hash function because it treats related DNS queries whose only difference is in their sub-domain (for example, 'time1.google.com' and 'time2.google.com'), as the same query. If we were to consider the entire URL, each member of this URL 'family' would hash to a different value and the dataset would not represent that these queries are related. This is particularly important as the sub-domain is often the part of the URL that changes most frequently and by ignoring it the model will be more robust to changes in device behavior.
On the other hand, considering only the top-level domain (TLD) would be too general, and completely unrelated domains would be hashed to the same value.

Each dataset is associated with a time delta value $t\Delta$ between 1s and 60s and is filtered to contain only the DNS queries whose timestamps fall between the DHCP timestamp and the sum of the DHCP timestamp and the time delta as shown in equation \ref{eq:time-delta}.

\begin{align} \label{eq:time-delta}
    \begin{split}
        &Dataset_{t\Delta} = \{Device, \{DNS_{hash},\ DNS_t\}\} \\
        s.t.:\ &DHCP_t < DNS_t <= (DHCP_t + t\Delta)
    \end{split}
\end{align}

Once the data is filtered by DNS timestamps, it is converted from storing $(DNS_{hash}, DNS_t)$ pairs to storing the DNS and its associated frequency. The frequency is calculated as the average number of times per second the $DNS_{hash}$ occurs between $DHCP_t$ and $(DHCP_t + t\Delta)$, see the final schema of the dataset in equation \ref{eq:freq}.

\begin{align} \label{eq:freq}
        Dataset_{t\Delta} = \{Device, \{DNS_{hash},\ DNS_{frequency}\}\}
\end{align}

In order to test how well the predictive models retain accuracy over time, we also generate and save datasets over a restricted set off dates, for example only containing experimental data captured between 2 days, rather than every experiment. This allows us to train models on data from particular dates and analyze its performance over other, unseen time periods.

\subsection{Data Pre-processing}
We split the dataset into features and labels, the features are normalized between 0 and 1 using a minmax scaler and the label categories are encoded using one hot encoding~\cite{potdar2017comparative}. 

This process must occur both for training and when using the models to make predictions on unseen data. For the purposes of training the model, we split the data into training and testing data in a $80:20$ ratio and we split the training data into training and validation data with the same ratio again.

\subsection{Neural Network Architecture}

We choose neural networks for the predictive model as they can learn complex behavior including both the presence of particular DNS queries and patterns that appear in the time domain. It is also possible to update a neural network's weights when more data becomes available~\cite{chen1999rapid} and different configurations of network can be readily compared~\cite{hunter2012selection}. We generate and compare neural networks across a design space of 4 parameters and hyperparameters shown in Table~ \ref{table:neural-network-params}, in order to find the optimal values.

\begin{table}[!bpt]
	\centering
	\rowcolors{3}{gray!10}{white}
	   \large
  \begin{tabular}{cc}
			\textbf{Parameter}& \textbf{Values tested}
			\\ \hline
\emph{Number of Hidden Layers} & 1,2,3
\\ \hline
\emph{Hash Resolution} & 4, 8, 16, 32, 64
\\ \hline
\emph{Time Delta} & 1 to 60
\\ \hline
\emph{Number of output Classes} & 27 and 30
\\ \hline
\end{tabular}

\caption{Parameters and hyperparameters for the neural networks under consideration.} 
\label{table:neural-network-params}
\end{table}

The dimension of the input layer of the neural network matches the hash resolution $h$ used to generate the dataset. This is because the input feature is the frequency with which each hashed domain is visited, and each hashed domain corresponds to one of the input neurons.

The remaining layers consisted of a varying number of dense hidden layers with 64 neurons followed by the output layer whose dimension is equal to the number of devices in the experiment (27 for manufacture granularity and 30 for device granularity). All layers use rectilinear activation except for the final layer which use softmax activation. The softmax activation $\sigma(z_i)$ is given by equation~\ref{eq:softmax} where $z$ is the input vector and $K$ is the number of classes.

\begin{figure}[t]
\centerline{\includegraphics[clip,width=0.9\columnwidth]{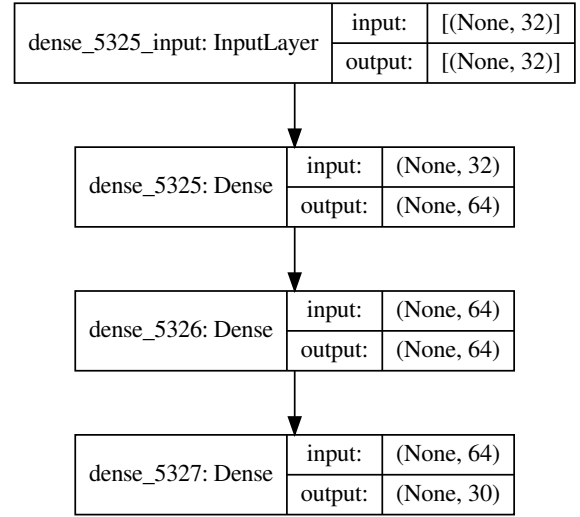}}
\caption{Example architecture of a neural network with 2 hidden layers and a hash resolution of 32.}
\label{fig:architecture}
\end{figure}

We choose categorical cross-entropy as the loss function as given by equation~\ref{eq:cross-entropy} where $N$ is the number of observations, $C$ is the number of categories and $p_{model}[y_i \in C_c]$ is the probability predicted by the model that observation $i$ belongs to class $c$. We chose categorical accuracy as the target metric and an Adam optimizer whose hyperparameters are shown in Table~\ref{tab:adam-params}. 
The architecture of an example neural network is shown in Figure~\ref{fig:architecture}.

\begin{table}[!bpt]
	\centering
	\rowcolors{3}{gray!10}{white}
	  \large	
	    \begin{tabular}{cc}
			\textbf{Hyperparameter}& \textbf{Value}
			\\ \hline
\emph{Learnig Rate} & 
0.001
\\ \hline
\emph{Beta 1} & 
0.9
\\ \hline
\emph{Beta 2} & 
0.999
\\ \hline
\emph{Epsilon} & 
1e-7
\\ \hline
\end{tabular}

\caption{Hyperparameters for the Adam optimizer.} 
\label{tab:adam-params}

\end{table}

\begin{align} \label{eq:softmax}
\sigma(z_i)=\frac{e^{x_i}}{\sum_{j=1}^{n}e^x_j}
\end{align}

\begin{align} \label{eq:cross-entropy}
-\frac{1}{N}\sum_{N}^{i=1}\sum_{C}^{c=1}1_{y_i\in C_c}log{\ p_{model}}[y_i \in C_c]
\end{align}

\subsection{Model Training}
\begin{figure}[t]
\centerline{\includegraphics[clip,width=1\columnwidth]{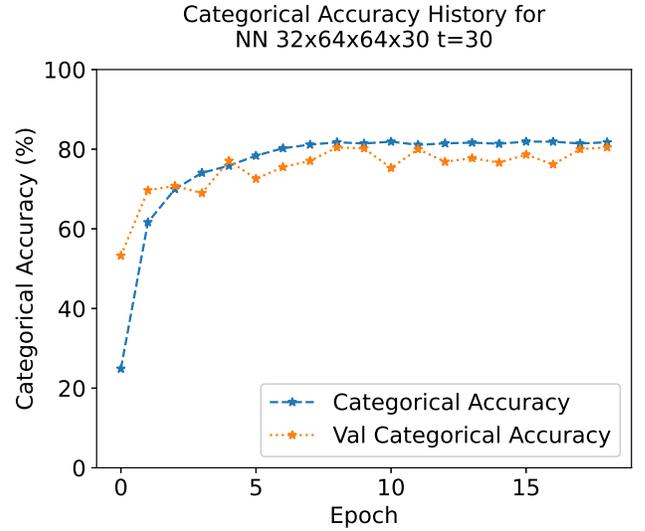}}
\caption{Categorical accuracy of a neural network with 2 hidden layers, a hash resolution of 32 and a time delta of 30 over number of epochs. Early stopping stops training at epoch 17.}
\label{fig:accuracy_history}
\end{figure}

We train neural networks with 1, 2 and 3 hidden layers for each hash resolution and time delta. We run the training over 100 epochs, however we implement early stopping to maximise the categorical accuracy and training would usually conclude at a much lower number of epochs. An example of the categorical accuracy training history for the training and validation data of a particular neural network is shown in Figure~\ref{fig:accuracy_history}. 
 
 When comparing large numbers of neural networks it is important to reduce the role that the random initialization of weights plays in the networks results. To accommodate this, we train each of the 1,800 neural network configurations four times with initialization weights initialized from different random seeds for reproducibility. We then average the accuracy of these four networks and use it for comparison between other network configurations. We then use the model with the highest accuracy to generate a macro f1 score which is the harmonic mean between precision and recall. 
 
 If $p_j$ and $r_j$ are the precision and recall for a given class $j$ within a set of classes $Q$, macro f1 score is given by equation~\ref{eq:macro-f1}.
 
 \begin{align} \label{eq:macro-f1}
    MacroF1 = \frac{1}{Q}\sum_{j = 1}^{Q} \frac{2 \cdot p_j \cdot r_j}{p_j + r_j}
 \end{align}
 
 Following training and evaluation, we collect each neural network object in a directory labelled with the time delta, hash resolution and number of hidden layers. The object acts as a wrapper for the 4 averaged models and contains the dataset and results of the best performing model, including the loss, categorical accuracy and macro f1 score. We save the best predictive model itself separately.

\section{Evaluation}\label{sec:evaluation}

  \begin{figure}[t]
\centerline{\includegraphics[clip,width=1\columnwidth]{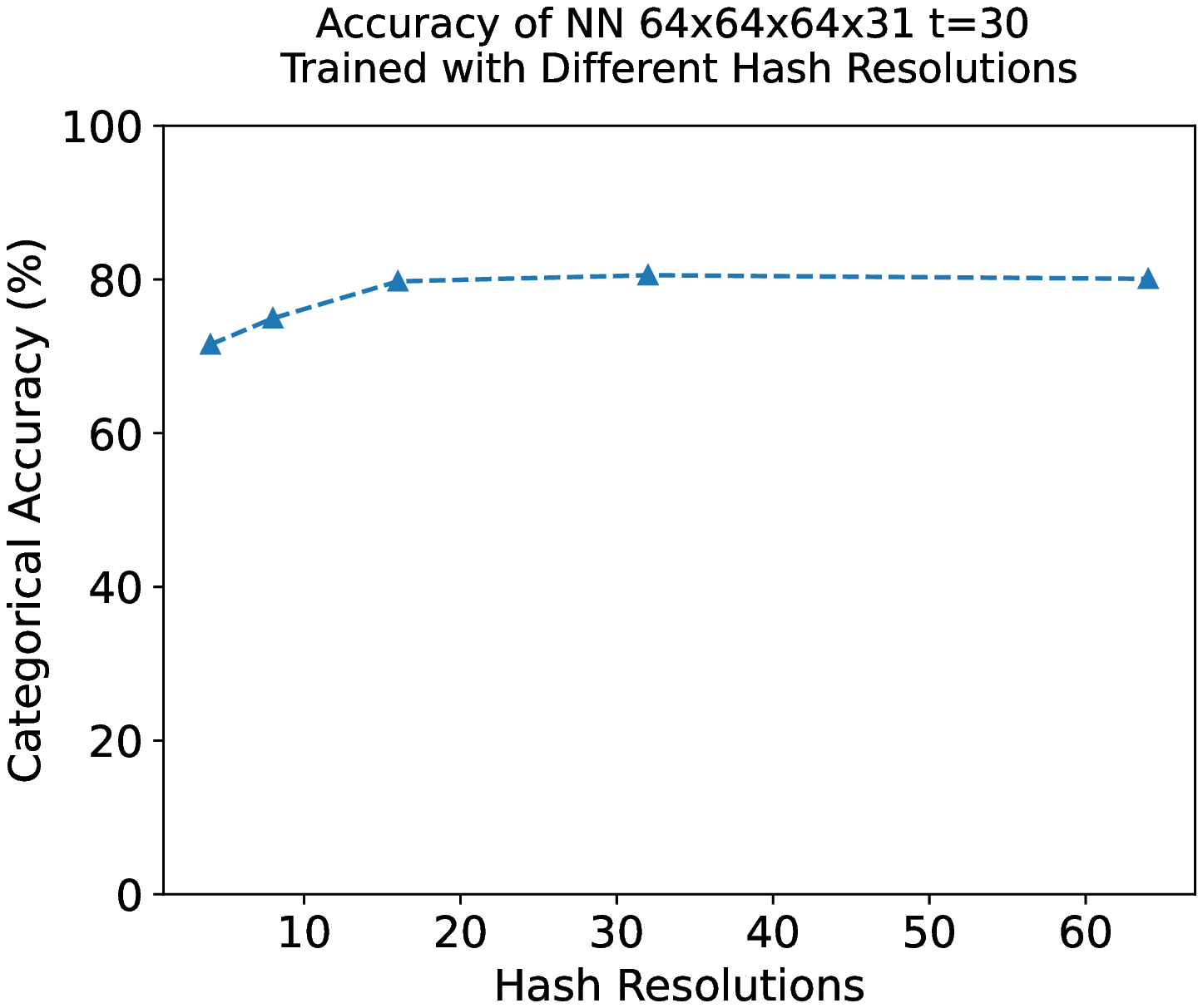}}
\vspace{0.3cm}
\caption{The categorical accuracy of she neural network with 2 hidden layers and a time delta of 30s. We train and test with 5 different hash resolutions to explore the effect of hash resolution value on model accuracy.}
\label{fig:hash-resolutions}
\end{figure}

\begin{figure}[t]
\centerline{\includegraphics[clip,width=1.1\columnwidth]{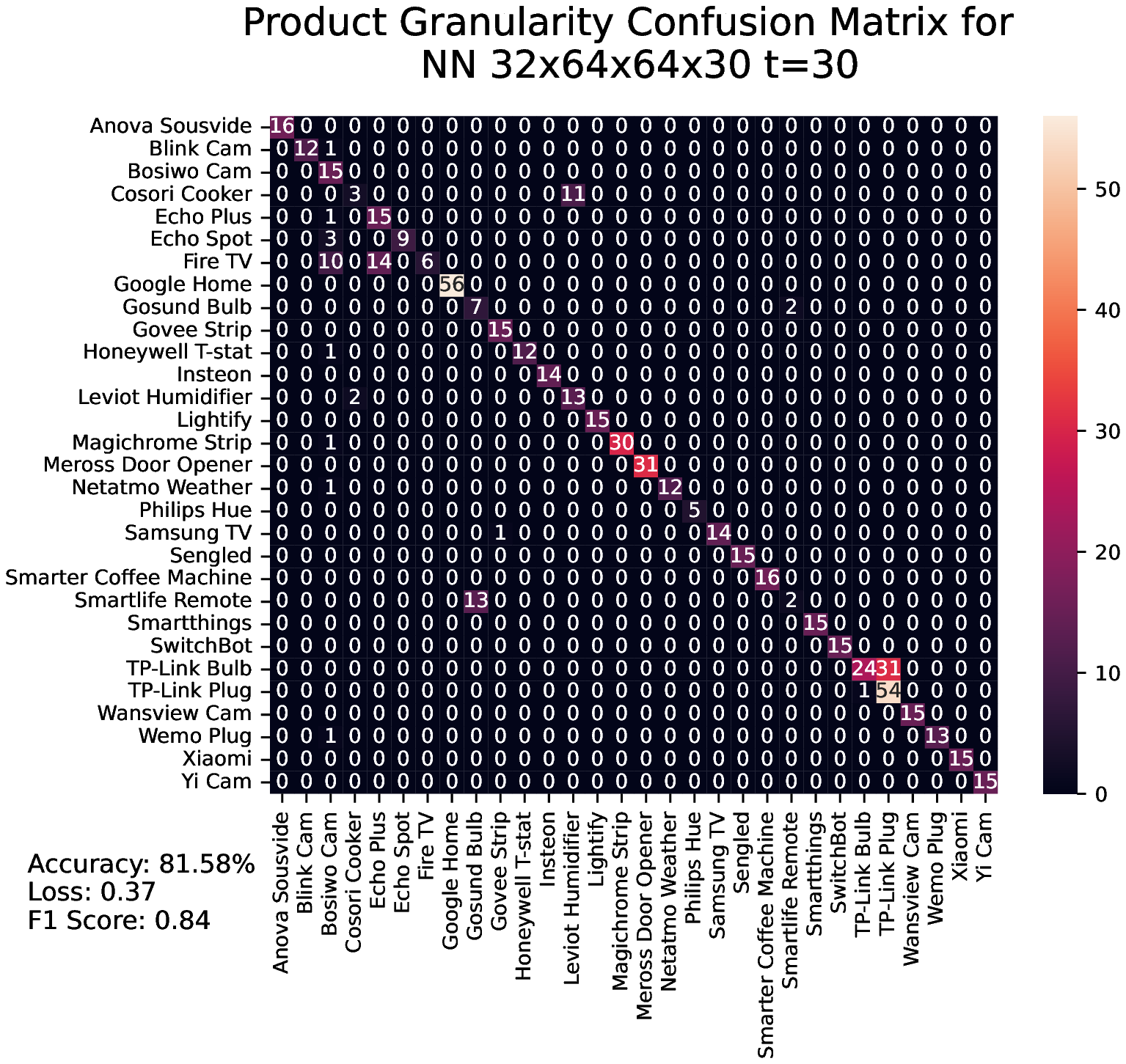}}
\caption{Product level granularity confusion matrix of a highly performing neural network with 2 hidden layers and a time delta of 30s and hash resolution of 32.}
\label{fig:product-confusion-matrix}
\end{figure}

\begin{figure}[t]
\centerline{\includegraphics[clip,width=1.1\columnwidth]{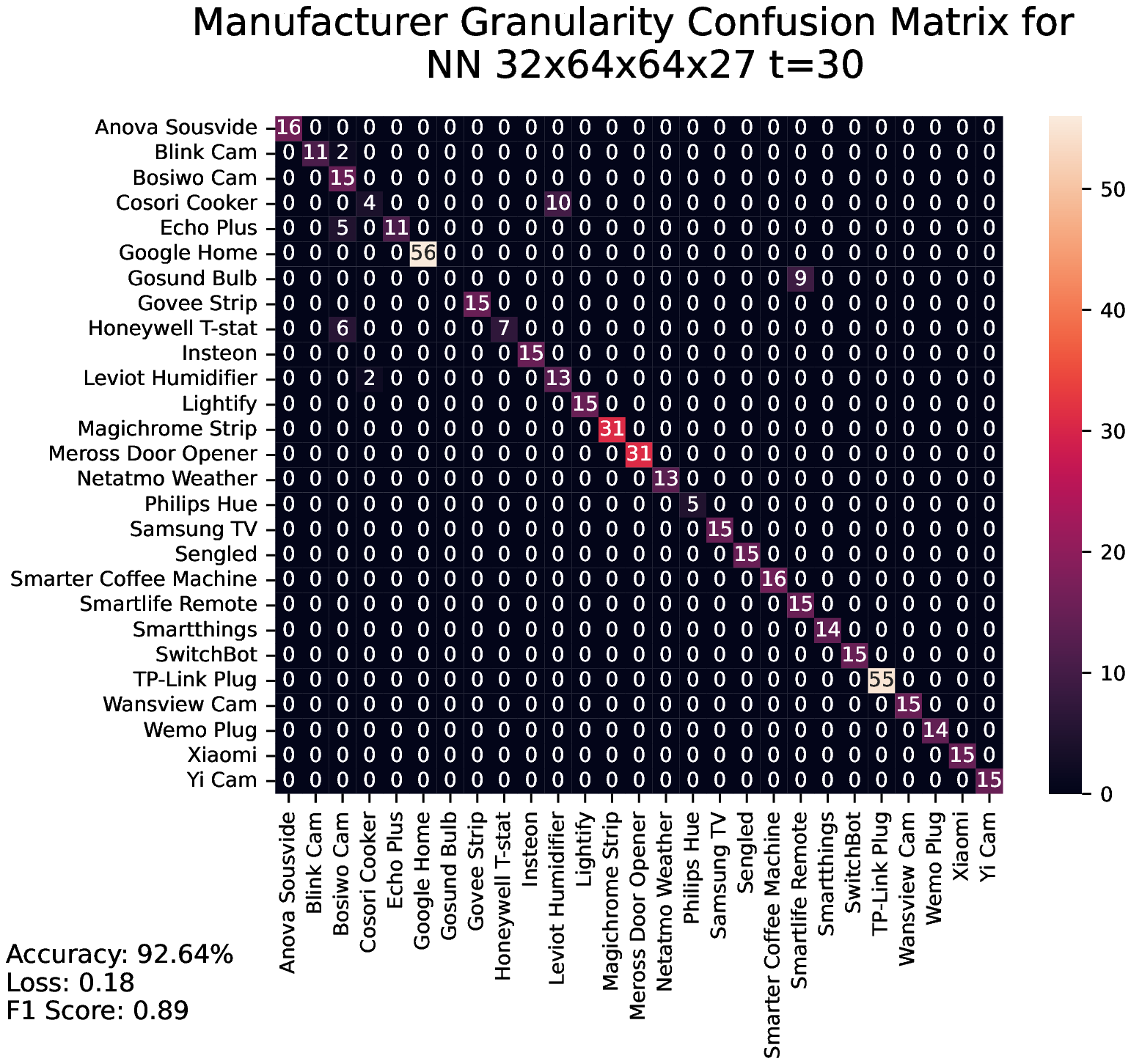}}
\caption{Manufacturer level granularity confusion matrix of a highly performing neural network with 2 hidden layers and a time delta of 30s and hash resolution of 32.}
\label{fig:manufacturer-confusion-matrix}
\end{figure}

In this section we evaluate the highest performing networks, and we compute the minimum amount of time necessary to reach maximum accuracy from the first few seconds of traffic after the device is connected.

\subsection{Highest Performing Networks}

By sorting the 1,800 models of different configurations by categorical accuracy, we can critically evaluate the neural network design space. The highest performing networks all have a hash resolution of at least 32. Figure~\ref{fig:hash-resolutions} shows that below 32 there is a reduction in accuracy and hash resolutions above 32 do not result in a significant increase in accuracy.

Results show that there is no notable increase in accuracy between neural networks with more than 2 hidden layers. This suggests that the function that classifies a device from its DNS traffic is not complex enough to require more than 2 hidden layers and that using more complex models may therefore increase the chance of over-fitting.

Figure~\ref{fig:product-confusion-matrix} shows the confusion matrix of a neural network trained on 30 seconds of data at product level granularity. Its categorical accuracy and macro f1 score are 82\%, and 0.84. 
Figure~\ref{fig:manufacturer-confusion-matrix} shows the confusion matrix for the same experiment conducted at manufacturer level granularity. Its categorical accuracy and macro f1 score are 93\% and 0.89. Both confusion matrices are constructed with the predicted device along the y-axis and the actual device along the x-axis. Classifications that do not lie on the main diagonal are misclassifications.

It can be seen that false positive identification occurs between the 2 TP-Link devices and between the 3 Amazon devices. Since these devices share a common manufacturer, they tend to query the same destinations which makes product level identification more difficult. 
The Smartlife~Remote, the Cosori~Cooker and the Gosund~bulb devices are the most frequently misclassified. By examining the dataset we conclude that this is because these 3 devices rarely make any DNS requests when compared to the other devices, so there is not enough data to make an accurate prediction. It may be necessary to measure DNS traffic over a longer time period before a reliable classification can be made for these devices.  

\subsection{Comparison of Different Time Deltas}

Once we establish the optimum neural network architecture and hash resolution, we consider the categorical accuracy of a network against the time deltas over which it is trained. The value of the time delta determines how many seconds of traffic the neural network is trained on.

\begin{figure}[t]
\centerline{\includegraphics[clip,width=1\columnwidth]{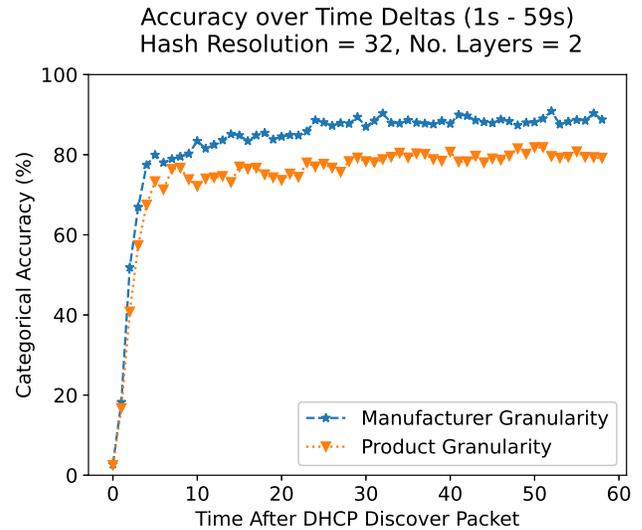}}
\caption{Accuracy against different time deltas. The neural network has a hash resolution of 32 and 2 hidden layers. The model's accuracy stops increasing significantly at around 30s, at both product and manufacturer granularity.}
\label{fig:accuracy-time}
\end{figure}

Figure~\ref{fig:accuracy-time} shows that the categorical accuracy of the neural network increases quickly between time deltas of 1 and 10 seconds. After this, the categorical accuracy increases at a slower rate until roughly 30 seconds, after which it does not increase significantly.
When making a prediction on new data, we now know that 30 seconds of DNS traffic should be captured in order to maximize the chances of accurate classification.

\subsection{Model Reliability Over Time}

To validate the model further it is important to test the behavior on unseen data. It is understood from the literature~\cite{kolcun2019case} that device identification models trained on data acquired through packet capture quickly become inaccurate as device behavior changes, for example if a device receives a software update from its manufacturer. We restrict the training data of the neural networks to data collected over 2 days and use the unseen data from the following days to test the networks performance. We test the method over different date ranges. 

\begin{figure}[t]
\centerline{\includegraphics[clip,width=0.9\columnwidth]{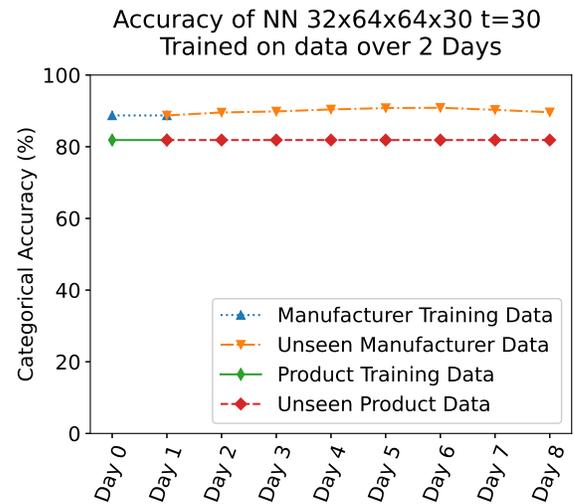}}
\caption{Plot of model accuracy over a week when testing on unseen data. A network trained on data for 2 days maintains its accuracy when tested on data from the following 7 days.}
\label{fig:train-test}
\end{figure}

Figure~\ref{fig:train-test} shows that even when a model is only trained on a single day's data the neural network maintains its high accuracy over the following week. This suggests that the DNS traffic of IoT devices does not change as frequently as the contents of the network packets. 

\section{Discussion}\label{sec:disc}

\subsection{Limitations}\label{sec:disc}

Our methodology has some limitations.
\\

\noindent \textbf{Model degradation.} Although we show that the model retains reliability over a week long time period, we can assume that there will be a point in time where the model's accuracy will degrade. It would be beneficial to test the accuracy degradation over a larger timescale in order to understand how often the model weights must be updated to retain accuracy. Past works have demonstrated that IoT DNS traffic is consistent over a period of at least 6 months~\cite{mandalari2021blocking}, which suggests the model may only require updating very rarely.
\\

\noindent \textbf{Scalability.} We demonstrate effective identification over a dataset of 30 devices with a relatively simple, 4 layer neural network. We do not yet understand if this model architecture will scale to a larger set of devices. It would also be useful to understand if the model could be used to accurately predict the presence of a non-IoT device. 
\\

\noindent \textbf{Unexpected behavior.} In very rare cases an IoT device may use hard-coded IP addresses rather than making DNS requests. The method of identification cannot be used for these types of devices.

\subsection{Future Directions}\label{sec:future-directions}

One of the possible solutions that we would like to investigate in the future is to use this method with on-device training to update the model locally to fit the latest DNS behavior. By measuring the DNS traffic of connected devices of known product and manufacturer, this data could be used to adjust the weights in the model and perpetually maintain a high accuracy. 

This methodology may be suitable for a crowdsourcing approach, where changes in model weights are shared between devices to benefit from information from a much larger dataset. This process would be privacy preserving by nature because all information about device traffic would be passed through the hashing function after which the original data could not be retrieved.

In order to support more fine-grained intra-manufacturer distinguishability, we aim to consider more seconds of initial traffic to improve the accuracy in these cases.

\section{Related work}\label{sec:related-work}

In recent years, a vast number of techniques for IoT device identification using machine learning have been investigated. 

The approach laid out by Meidan \etal~\cite{meidan2017profiliot} is able to predict a device's brand and model with 99\% accuracy. Their approach firstly distinguishes IoT devices from non IoT devices by examining the user agent property of the HTTP header from captured PCAP files. The second session identifies the type of IoT device through logical characteristics of the captured packets. 
Similarly, Aksoy \etal's method~\cite{aksoy2019automated} achieves device classification with an accuracy of over 95\% through analysis of a single captured packet. They also observed that when devices share a manufacturer the accuracy of the classification is reduced. 

A different approach from Kotak and Elovici \cite{10.1007/978-3-030-57805-3_8} achieves IoT device identification of 99\% also using deep neural networks. The network makes a prediction based on patterns found in a device's traffic; the raw binary PCAP file is truncated and converted into 28 x 28 greyscale images which are used as input to a simple convolutional neural network. This approach is efficient in that it utilizes the PCAP file in its raw form and does not need to explicitly extract features. 

The methodology used by Kolcun \etal~\cite{kolcun2019case} explores various predictive models including a decision tree classifier, random forest classifiers and 3 different neural networks. The data features consist of various statistical properties from PCAP files, for example kurtosis and skewness of packet size and inter-packet gap, source and destination ports and domains. The results show the method with the best accuracy was the random forest classifier, however the methods do not retain their accuracy over longer periods of time. 
All the aforementioned techniques use packet capture to extract features from an IoT device's traffic, while our aim is to use only DNS log traffic. 

The approach from Perdisci \etal~\cite{9230403} however, resembles our method of feature extraction as it uses the query URLs of a device's DNS requests. This approach uses the whole URL rather than the SLD and uses a naive document retrieval algorithm to match URLS to devices. By building on top of this methodology with advanced machine learning techniques, we are able to learn behavior in the time domain in addition to matching DNS queries and achieve high accuracy over a larger set of devices, without requiring full packet capture. Moreover, Perdisci \etal approach set the time window length to be one hour, whereas our methodology uses only the first 30 seconds of DNS traffic.

\section{Conclusion}\label{sec:conclusion}

Consumer IoT devices are already very popular, and their usage is expected to grow further. There is a need to track their deployment without deep packet inspection or active measurements, both intrusive and unscalable methods for large deployments on a device at the edge. Our insight is that many IoT devices contact a small number of domains and it is possible to detect such devices at scale from sampled DNS measurements following a devices first boot.

Our method is able to detect 30 IoT products at manufacturer granularity with 93\% accuracy and at product granularity with 82\% accuracy. While this detection may be useful to perform device-specific DNS filtering of IoT devices at home, it raises concerns about the general detectability of such devices and the corresponding human activity.  

We have also established that roughly 30 seconds of IoT device DNS traffic should be observed in order to maximize the accuracy of the prediction and that models can retain this accuracy over the course of a week. We show that product level identification is highly accurate but may mis-classify devices that share a manufacturer. Conducting the experiment at the manufacturer level eliminates this confusion and can classify nearly every manufacturer accurately.

\section*{Acknowledgments}

We thank the anonymous reviewers for their constructive feedback. The research in this paper was partially supported by the EPSRC (Databox EP/N028260/1, DADA EP/R03351X/1, HDI EP/R045178/1, and Impact Acceleration Account (IAA)).

\balance
\bibliographystyle{unsrt}
\interlinepenalty=10000
\bibliography{paper}

\end{document}